\documentclass{article} 
\usepackage{iclr2026_conference,times}

\usepackage{amsmath,amsfonts,bm}

\def\eqref#1{equation~\ref{#1}}

\def\1{\bm{1}}

\DeclareMathAlphabet{\mathsfit}{\encodingdefault}{\sfdefault}{m}{sl}
\SetMathAlphabet{\mathsfit}{bold}{\encodingdefault}{\sfdefault}{bx}{n}

\usepackage{hyperref}
\usepackage{url}
\usepackage{subcaption}
\usepackage{lipsum}
\usepackage{tikz}
\usepackage{amsmath}
\usepackage{subcaption}
\usepackage{booktabs}
\usepackage{siunitx}
\usepackage{array}
\usepackage[ruled,vlined]{algorithm2e}
\usepackage{wrapfig} 
\usetikzlibrary{arrows.meta, positioning}
\usetikzlibrary{shapes.geometric}
\usetikzlibrary{calc}
\usetikzlibrary{fit,backgrounds}

\definecolor{offblue}{RGB}{0,128,255}

\tikzset{
  block/.style = {
    rectangle, 
    rounded corners=4pt,
    draw,
    very thick,
    minimum width=1cm,
    minimum height=1cm,
    align=center,
    fill=gray!6
  },
  arrow/.style = {
    -{Stealth[length=6pt,width=6pt]},
    line width=1.6pt,
    shorten >=3pt,
    shorten <=3pt,
  },
  arrow_inference/.style = {
      -{Stealth[length=6pt,width=6pt]},
  line width=1.6pt,
  shorten >=1pt,
  shorten <=1pt,
  },
  arrow_training/.style = {
    -{Stealth[length=6pt,width=6pt]},
  line width=1.6pt,
  shorten >=1pt,
  shorten <=1pt,
  },
}

\title{Learning to Rank the Initial Branching Order of SAT Solvers}

\author{%
  Arvid Eriksson\thanks{Corresponding author: \texttt{arveri@kth.se}}$^{\,\,\,, 1}$,
  Gabriel Poesia$^2$,
  Roman Bresson$^3$,
  Karl H. Johansson$^1$,
  David Broman$^1$ \\
  \vspace{0.15cm} \\
  $^1$KTH Royal Institute of Technology \\
  $^2$Kempner Institute at Harvard University \\
  $^3$Mohamed Bin Zayed University of Artificial Intelligence
}

\iclrfinalcopy 
\begin{document}

\maketitle

\begin{abstract}




Finding good branching orders is key to solving SAT problems efficiently, but finding such branching orders is a difficult problem. Using a learning-based approach to predict a good branching order before solving, therefore, has potential. In this paper, we investigate predicting branching orders using graph neural networks as a preprocessing step to conflict-driven clause learning (CDCL) SAT solvers. We show that there are significant gains to be made in existing CDCL SAT solvers by providing a good initial branching. Further, we provide three labeling methods to find such initial branching orders in a tractable way. Finally, we train a graph neural network to predict these branching orders and show through our evaluations that a GNN-initialized ordering yields significant speedups on random 3-CNF and pseudo-industrial benchmarks, with generalization capabilities to instances much larger than the training set. However, we also find that the predictions fail to speed up problems with a large number of variables, and more difficult industrial instances. We attribute this to the solver's dynamic heuristics, which rapidly overwrite the provided initialization, and to the complexity of these instances, making GNN prediction hard.

\end{abstract}

\section{Introduction}

The Boolean satisfiability problem (SAT) is the archetypal NP-complete problem, to which all NP problems can be reduced \citep{biere_handbook_2009}. These types of problems are central in formal verification and automated theorem proving, and solving them efficiently is therefore a key step towards automating proofs. State-of-the-art SAT solvers rely on several heuristics to tame the combinatorial search space inherent to the problem. While they often work well on average, these fixed, handcrafted heuristics still fail to effectively guide search in a significant number of cases: many problem instances of practical value are still out of reach even for the best existing SAT solvers.


Rather than relying on static rules, an alternative route is to learn heuristics from data. In particular, graph neural networks (GNNs) can learn non-trivial structure about SAT instances \citep{selsam_learning_2018}, and reinforcement learning can be used to learn effective guiding heuristics for SAT solvers \citep{kurin_can_2020}. However, neural network inference is computationally expensive, and using neural heuristics \emph{during} search tends to underperform using simpler heuristics that are orders of magnitude faster, even if neural heuristics might make better decisions on average.

An important open problem is therefore developing efficient hybrid solvers that combine the capacity of neural networks with the efficiency of rule-based state-of-the-art SAT solvers. The idea is to use a neural network to guide the solver in key decisions while relying on the rule-based heuristics for pure efficiency. Some such hybrid solvers have been proposed, such as RDC-SAT \citep{zhai_learning_2025}, a divide and conquer deep reinforcement learning hybrid solver, and NeuroCore \citep{selsam_guiding_2019} for UNSAT core prediction that both occasionally poll a GNN for some key decision. These works, however, require a heavily modified solver that needs to communicate with a GNN during solving. A less invasive approach is NeuroBack  \citep{wang_neuroback_2024-1}, which provides the solver with an initial guess of the assignment for each variable. However, this only targets one of the two main decisions in a SAT solver, phase selection. The remaining key decision --- which variable to branch on at each point --- remains unchanged.

\begin{figure}[]
    \centering
    \smaller
\begin{subfigure}{\textwidth}
\centering
\begin{tikzpicture}[node distance=0.45cm and 0.45cm]

  \node[block, fill=black!10, draw=black!10] (Input) {\textbf{SAT instance} \\
  $\phi \in \Phi$};
    \node[block, right=of Input, fill=orange!30, draw=orange!80, dotted] (Label) {\textbf{Label instance} \\ $\boldsymbol y = f_{label}(\phi)$};

  \node[block, right=of Label, fill=black!10, draw=black!10] (Convert) {\textbf{Graph conversion} \\ $\boldsymbol x := f_{graph}(\phi)$};

  \node[block, right=of Convert, fill=purple!20, draw=purple!80, dashed] (GNN) {\textbf{GNN} \\ $\boldsymbol s := f_\theta( \boldsymbol x)$};

  \node[block, right=of GNN, fill=black!10, draw=black!10] (Loss) {\textbf{Ranking loss} \\ Minimize $\mathcal{L}(\boldsymbol s, \boldsymbol y)$};

\node[draw=black!50, solid, thick, fit=(Convert), inner sep=1mm,
      label={[fill=white, yshift=-1mm]above:\textbf{Section \ref{sec:graph}}}] {};

\node[draw=black!50, solid, thick, fit=(GNN)(Loss), inner sep=1mm,
      label={[fill=white, yshift=-1mm]above:\textbf{Section \ref{sec:architecture}}}] {};

\node[draw=black!50, solid, thick, fit=(Label), inner sep=1mm,
      label={[fill=white, yshift=-1mm]above:\textbf{Section \ref{sec:labeling}}}] {};

\draw[arrow_training] (Input) -- (Label);
  \draw[arrow_training] (Label) -- (Convert);
  \draw[arrow_training] (Convert) -- (GNN); 
  \draw[arrow_training] ([yshift=2mm]GNN.east) -- ([yshift=2mm]Loss.west);  
  \draw[arrow_training] ([yshift=-2mm]Loss.west) -- ([yshift=-2mm]GNN.east);
\end{tikzpicture}
\caption{Training}
\label{fig:training}
\end{subfigure}

\begin{subfigure}{\textwidth}
\centering
    \begin{tikzpicture}[node distance=0.45cm and 0.45cm]
  
  \node[block, fill=black!10, draw=black!10] (Input) {\textbf{SAT instance} \\ $\phi = (x_1 \vee \neg x_2)$ \\ $\wedge (x_2 \vee x_3)$};

  \node[block, right=of Input, fill=black!10, draw=black!10] (Convert) {\textbf{Graph conversion} \\ $\boldsymbol x := f_{graph}(\phi)$};
  
  \node[block, right=of Convert, fill=purple!20, draw=purple!80, dashed] (GNN) {\textbf{GNN} \\ $\boldsymbol s := f_\theta( \boldsymbol x)$};

  \node[block, right=of GNN, fill=black!10, draw=black!10] (Scores) {\textbf{Scores}
  \\ $s_1 = 0.4$ \\ $s_2 = 0.7$ \\ $s_3 = 0.1$};
  \node[block, right=of Scores, fill=black!10, draw=black!10] (Output) {\textbf{Initial order} \\ $[x_2,x_1,x_3]$};
  \node[block, right=of Output, fill=orange!30, draw=orange!80, dotted] (Solver) {\textbf{SAT solver} \\ SAT / UNSAT};

\node[draw=black!50, solid, thick, fit=(Input)(Solver)(Scores), inner sep=1mm,
      label={[fill=white, yshift=-2mm]above:\textbf{Section \ref{sec:experiments}}}] {};

  \draw[arrow_inference] (Input) -- (Convert);
  \draw[arrow_inference] (Convert) -- (GNN);
\draw[arrow_inference] (GNN) -- (Scores);
  \draw[arrow_inference] (Scores) -- (Output);
  \draw[arrow_inference] (Output) -- (Solver);
\end{tikzpicture}

\caption{Inference}
\end{subfigure}

    \caption{Pipeline for training (top) and inference (bottom) for a branch order-predicting GNN preprocessing step. The orange boxes (dotted outline) signify steps that use a modified SAT solver, while purple boxes (dashed outline) signify steps using a graph neural network.}
    \label{fig:inference}
\end{figure}
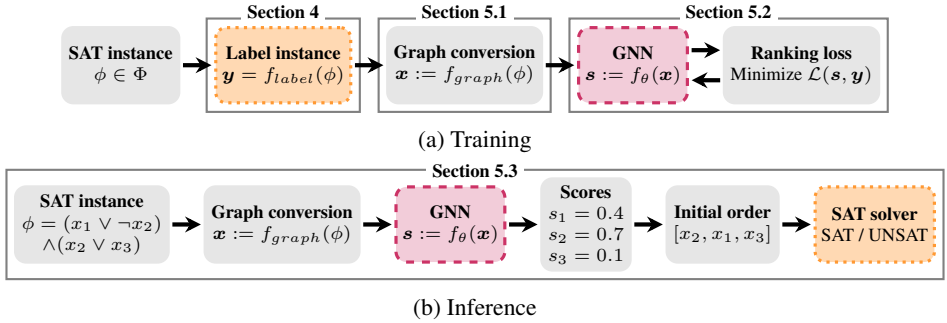

In this paper, we investigate predicting the initial branching order as a way to speed up existing backtracking SAT solvers. We make the following contributions:


\begin{itemize}
    \item We experimentally verify that the initial branching order has a significant impact on the overall solving time and show the possible gains of a good order initialization (Section \ref{sec:branchtime}).
    \item We propose three heuristics for identifying fast learnable branching orders, to be used as labels for a predictive GNN (Section \ref{sec:labeling}).
    \item We show that a GNN pipeline (Figure \ref{fig:inference}) can predict such a branching order with speed-ups over multiple domains (Section \ref{sec:neurobranch}), and often generalizes to instances that are up to an order of magnitude larger than instances given during training.
\end{itemize}


Our experience across two SAT solvers, MiniSat \citep{sorensson_minisat_2005} and CaDiCaL \citep{biere_cadical_2024}, demonstrates that predicting the initial branching order is itself a complex problem impacted by many design choices. Our initial experiments show that there are significant gains to be made in predicting a good initial branching order in generated SAT instances. However, we also observe and describe major challenges in obtaining similar speed-ups in industrial instances, which pose their own unique challenges for training due to their size and structure. Overall, our work outlines several potential lines for future work by further exploring the design space for hybrid SAT solvers, identifying the variable branching order as an underexplored but impactful meeting point for neural networks and state-of-the-art solvers.

\section{Related Work}

\textbf{End-to-end neural SAT solvers.}
Using neural networks to solve SAT instances is a recent trend where NeuroSAT \citep{selsam_learning_2018} was a seminal work. NeuroSAT sees SAT solving as a binary classification problem where a GNN is used to predict whether the instance is satisfiable or not. Multiple works, such as SATformer \citep{shi_satformer_2023} and SAT-GATv2 \citep{chang_sat-gatv2_2025} push the same idea further with higher predictive accuracy. However, the predictive nature of these works does not provide any solid guarantees on the solution, and their accuracy quickly drops for larger instances.

\textbf{Hybrid solvers with online neural heuristics.}
One way to guarantee a solution while still utilizing the predictive capabilities of neural networks is through a hybrid solver, a standard search-based SAT solver that is enhanced with some type of prediction from a neural network. Examples include NeuroCore \citep{selsam_guiding_2019}, which integrates a GNN into an existing solver to continuously overwrite the branching order toward regions predicted to contain an unsatisfiable core, thus speeding up solving while having all the guarantees that come with a SAT solver. Graph-Q-SAT \citep{kurin_can_2020} and the GNN proposed in \cite{yolcu_learning_2019} similarly polls a GNN, but instead at every step, and use reinforcement learning to learn an optimal branching heuristic. Likewise, RDC-SAT \citep{zhai_learning_2025} polls a GNN every time a split is done in parallel SAT solving. However, all of the above methods require frequent polling of a GNN, which is computationally expensive for large instances, whereas most classical solver heuristics run in constant time. 

\textbf{Hybrid solvers with neural initialization heuristics.}
A good compromise is calling the GNN a single time before solving and using it to initialize the internal structures of the solver, thus providing a warm start. NLocalSAT \citep{zhang_nlocalsat_2020} applies this to stochastic local search solvers by predicting an initial solution, granting a suitable initial guess. NeuroBack \citep{wang_neuroback_2024-1} adapts this strategy to conflict-driven clause learning (CDCL) solvers by predicting the initial phase (polarity) for each variable. Notably, NeuroBack is one of the few learning-based approaches to demonstrate wall-clock speedups for state-of-the-art solvers on large industrial instances. However, phase selection is only one of the two key parts of a backtracking solver, the other being branch selection. Our work, therefore, is the first to investigate how suitable initial branching orders can be predicted for backtracking SAT solvers.





\section{Branching Order's Impact on Solving Time}
\label{sec:branchtime}

The branching order in backtracking SAT solvers is known to be important for solving efficiently \citep{nordstrom_interplay_2015-1}. However, to the best of our knowledge, the magnitude of this effect has not been precisely characterized yet. To start, we now empirically test the impact of the solving order by quantifying the solving run time when starting to branch on a specific variable.

We perform the following experiment using CaDiCaL \citep{biere_cadical_2024}: We randomly sample four generated 3-CNF uniformly random (of a specific number of variables) and four industrial satisfiable SAT instances from SAT competitions from 2015 to 2024 that CaDiCaL solves in less than 10 seconds. For these instances we sample 50 random variables and force each sampled variable to be first 1000 times, with the order of the remaining variables chosen at random. This approach allows us to quantify how a one-variable suggested branching order can contribute to the solving time. 

\begin{figure}[htbp]
    \centering
    \includegraphics[width=\linewidth]{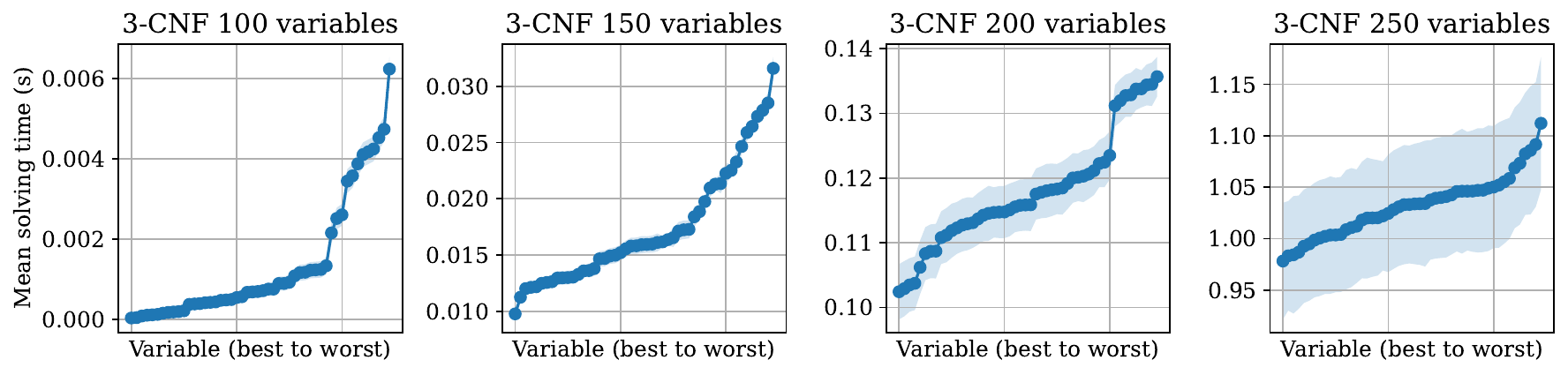}
    \includegraphics[width=\linewidth]{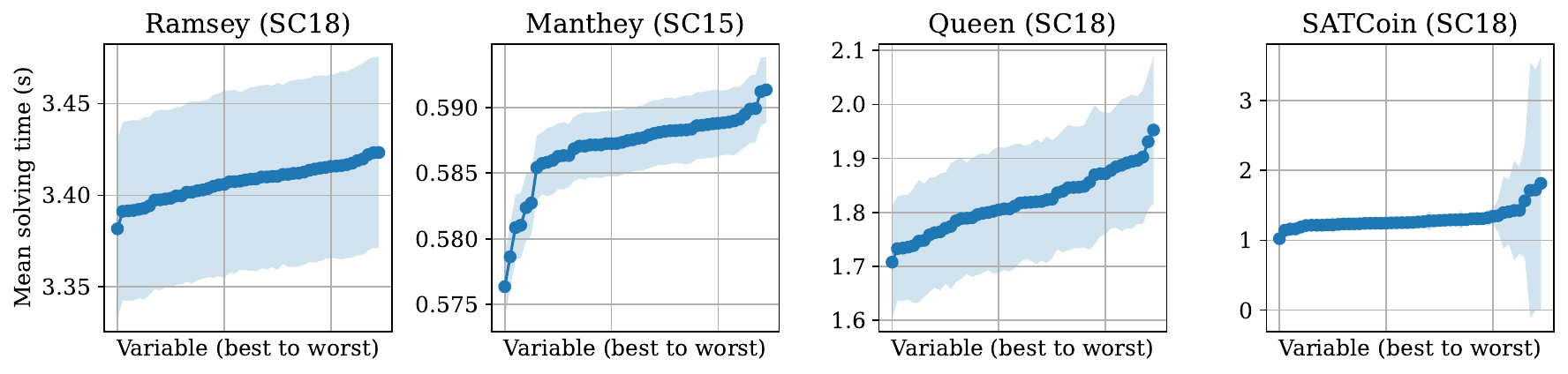}

    \caption{Mean solving time for setting one of 50 randomly sampled variables first, with the remaining order randomized. The shaded areas indicate a 95\% confidence interval over 1000 iterations for each variable.}
    \label{fig:motivational}
\end{figure}

In Figure \ref{fig:motivational} and Table \ref{tab:branch}, we see that the first variable chosen significantly impacts the solver's solving time for most instances. This illustrates the potential effect that a good initial branching order can have and suggests that there is an underlying ranking of variables from good to bad to branch on early. The low variance in some instances further suggests that the resulting solving time is a result of the variable rather than the random order that succeeds it, signaling some type of variable feature that is learnable by a GNN.

The results further show that there is a potential speedup (roughly 1\% to 10\% for industrial instances) from choosing a good first variable to branch on compared to the median. Such an effect would be further amplified if the full branch order (not just the first variable) is chosen suitably, thus motivating the remainder of this paper.


\begin{table}[t]
\centering
\caption{Solving time and speedup when choosing the best and 10th-percentile variables, respectively, compared to the median, when forcing initial branching to start on a specific variable.}
\label{tab:branch}

\setlength{\tabcolsep}{3.5pt} 

\begin{tabular}{
    l 
    S[table-format=6.0, group-separator={\,}] 
    S[table-format=6.0, group-separator={\,}] 
    S[table-format=1.4] 
    S[table-format=4.1] 
    S[table-format=3.1]
}
\toprule
\textbf{Instance} & {\textbf{Variables}} & {\textbf{Clauses}} & {\textbf{Median (s)}} & {\textbf{Best speedup (\%)}} & {\textbf{Top 10\% speedup (\%)}} \\
\midrule
3-CNF 100 & 100    & 430    & 0.0007 & 2233.3 & 536.4 \\
3-CNF 150 & 150    & 645    & 0.016  & 63.1   & 30.8  \\
3-CNF 200 & 200    & 860    & 0.12   & 13.1   & 9.1   \\
3-CNF 250 & 250    & 1075   & 0.98   & 5.6    & 4.1   \\
Ramsey    & 37410  & 162400 & 3.41   & 0.8    & 0.5   \\
Manthey   & 3996   & 29400  & 0.59   & 2.0    & 0.9   \\
Queen     & 256    & 6744   & 1.82   & 6.4    & 4.6   \\
SATCoin   & 134781 & 648806 & 1.26   & 22.5   & 5.4   \\
\bottomrule
\end{tabular}
\end{table}

\section{Identifying Good Branching Orders}
\label{sec:labeling}


In order to train a graph neural network to predict the best branching order for a given problem, we first need an estimate of the \textit{best} order for said problem. To do this, we can either use information directly from the solver collected while solving the problem (white box) or treat the solver as a black box and optimize the solving time of the problem over branching orders. We suggest three labeling methods, conflict labeling, first variable labeling, and optimized labeling. We further compare them in terms of solver speed-up when used as labels for a GNN in Section \ref{sec:neurobranch}. Pseudocode for first variable and optimized labeling is provided in Appendix \ref{app:labelapp}.

\subsection{Conflict Labeling}
For the white box approach, we can label each problem by ranking each variable according to the total number of conflicts it participated in during solving. This yields a simple labeling that only requires solving a problem once to label it and initally branch on conflict-prone variables. This builds on the assumption that beginning to branch on the most conflict-prone variables speeds up solving. Such an assumption is justified, however, since high-conflict variables are likely backdoor variables, whose early resolution makes the problem trivial \citep{williams_backdoors_2003}. 


\subsection{First Variable Labeling}
First variable labeling assigns each variable a score by averaging the number of propagations over multiple solver runs where that variable is forced first, and the remaining order is randomized (similar to the experiment in Section \ref{sec:branchtime}). The label is the variables sorted by these scores. This gives a labeling method that builds on the individual contributions of each variable, which ideally is easy for a GNN to learn since it attempts to isolate each variable's individual contribution while negating the noise provided by the random remaining order.

\subsection{Genetic Labeling}
Genetic labeling instead frames labeling as an optimization problem: find the variable order that minimizes the number of propagations. Starting from $k$ random permutations of the variables, we keep the best permutation and iteratively improve it over $m$ generations by generating new candidates by randomly swapping variables in the permutation, in a hill-climbing fashion. 



\subsection{Using Predicted Branching Orders in the Solver}
Using the labeling strategies above, our ultimate goal is to label a large set of training instances and learn to predict a good variable branching order for new problems at inference time. However, most CDCL solvers do not have a built-in interface for providing them with a suggested initial branching order. Thus, to enable our experiments, a key design decision lies in how to communicate our predicted order to the solver. Solvers tend to keep an ``activity level'' associated with each variable, typically updated as variables are found to be part of conflicts. We find these activity levels to be a simple way to initialize variables with an ordering. This observation lets us modify the two SAT solvers that we use here: MiniSat \citep{sorensson_minisat_2005} and CaDiCaL \citep{biere_cadical_2024}. In both these solvers, our initial order is integrated seamlessly into the solver by initializing the activity levels used in the VSIDS \citep{moskewicz_chaff_2001} heuristic, which determines which variable to branch on. For CaDiCaL, the VMTF \citep{ryan_efficient_2004-1} queue is also initialized to the suggested order.

\section{Predicting Branching Order}
\label{sec:neurobranch}


In this section, we investigate how well a GNN can learn the branching order heuristics suggested in Section \ref{sec:labeling} and how well those predictions speed up the overall solving time of various types of SAT random, pseudo-industrial, and industrial instances.

\subsection{Graph Representation}
\label{sec:graph}

\begin{wrapfigure}[11]{r}{0.25\textwidth} 
    \includegraphics[width=\linewidth]{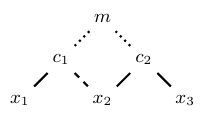}
    \caption{Graph of \\ $(x_1 \vee \neg x_2)\wedge(x_2 \vee x_3)$}
    \label{fig:graph}
\end{wrapfigure}

Before using the GNN, we need to convert the SAT instances to graphs (see Figure \ref{fig:training}). A SAT instance can be converted to a graph as illustrated by the example in Figure \ref{fig:graph}. The SAT instances are represented as in \cite{yolcu_learning_2019} as tripartite graphs where variables ($x_i$) and clauses ($c_i$) are represented as nodes, and each edge signifies that a variable is in a clause. Each edge also has a weight signifying whether the variable occurs negated (solid) or not (dashed) in the clause. We also add a meta node ($m$) connected to every clause node for quicker message passing as done in \cite{wang_neuroback_2024-1}. Finally, we enhance the features of each node with explicit one-hot encodings for which quartile the node's degree belongs to in comparison to all other nodes of the same type, as this slightly increases performance.

\subsection{Graph Neural Network and Loss Function}
\label{sec:architecture}

We use a GNN $f_\theta$ that takes as input a SAT instance $\boldsymbol x$ represented as a graph and outputs $\boldsymbol s := f_\theta(\boldsymbol x)$, which contains a score for every variable in the SAT instance. The predicted order is then retrieved by sorting the variables by their predicted score. Our implementation uses the GNN architecture from NeuroBack \citep{wang_neuroback_2024-1}, as it allows for attention mechanisms while keeping a linear memory complexity.

Since we want the GNN to predict an optimal order, we cast this as a ranking problem, where our goal is to match the variable's rank in the reference order generated when labeling training instances. To that end, we can use tools from the field of information retrieval, specifically loss functions used in learning to rank \citep{liu_learning_2007}. 
We use LambdaRank \citep{burges_learning_2005} since it emphasizes correctly placing variables with high \textit{relevance} near the top of the predicted order. For a variable $x_i$, we define its relevance as $rel_{x_i} = 1/log_2(R(x_i))$ where $R(x_i)$ denotes the rank of $x_i$ in the ground-truth ordering of the corresponding SAT instance. This definition of relevance reflects our priorities, as the most important variables will be assigned very high relevance, while variables beyond the top are less important. This also aligns with the tree-like search of the SAT solver, where the first decisions have a disproportionate impact on performance. 


\subsection{Evaluation}
\label{sec:experiments}

To evaluate our approach, we train branching predicting GNNs and evaluate them using MiniSat \citep{sorensson_minisat_2005} and CaDiCaL \citep{biere_cadical_2024}. We train and evaluate on three different types of distributions: random 3-CNF SAT, G4SATBench, and SAT competition from 2015 to 2024.

When training and evaluating we use the number of propagations as our performance metric, as it is proportional to solving time while being system independent and more reliable for small instances. We label and evaluate on the CPU and train on the GPU, see Appendix \ref{app:setup} for the exact setup and time spent. Running the GNN on the CPU also equates to an additional cost, which is negligible in comparison to the solving time for instances with more than 150 variables, but otherwise significant. See Appendix \ref{app:inference} for a comparison of inference and runtime and Appendix \ref{app:hyperparams} for hyperparameters used.

First, we investigate the GNN-enhanced solvers' performance on random 3-CNF SAT instances. We generate instances using a variable-clause ratio of 4.5, corresponding to the phase transition threshold where the frequency of SAT and UNSAT instances is approximately the same and the difficulty to solve the highest \citep{crawford_experimental_1996}. These samples are equally distributed between 20 and 40 variables, and SAT and UNSAT. We choose to train on smaller instances as the results in Section \ref{sec:branchtime} signal that labels for smaller instances have less variance and therefore are easier to learn. We train on 9000 instances and validate on 1\,000 instances. For testing, we use the full Uniform Random-3-SAT SATLIB benchmark suite \citep{hoos_satlib_2000} that includes similar problems to those trained on in the variable ranges of 20 to 250 variables. For each number of variables, there are 100 SAT and 100 UNSAT instances (1000 for 50 and 100 variables, and none for UNSAT at 20 variables). We train one GNN per labeling method and solver, thus resulting in six different GNNs.

\begin{figure}[ht]
    \centering
    \includegraphics[width=\linewidth]{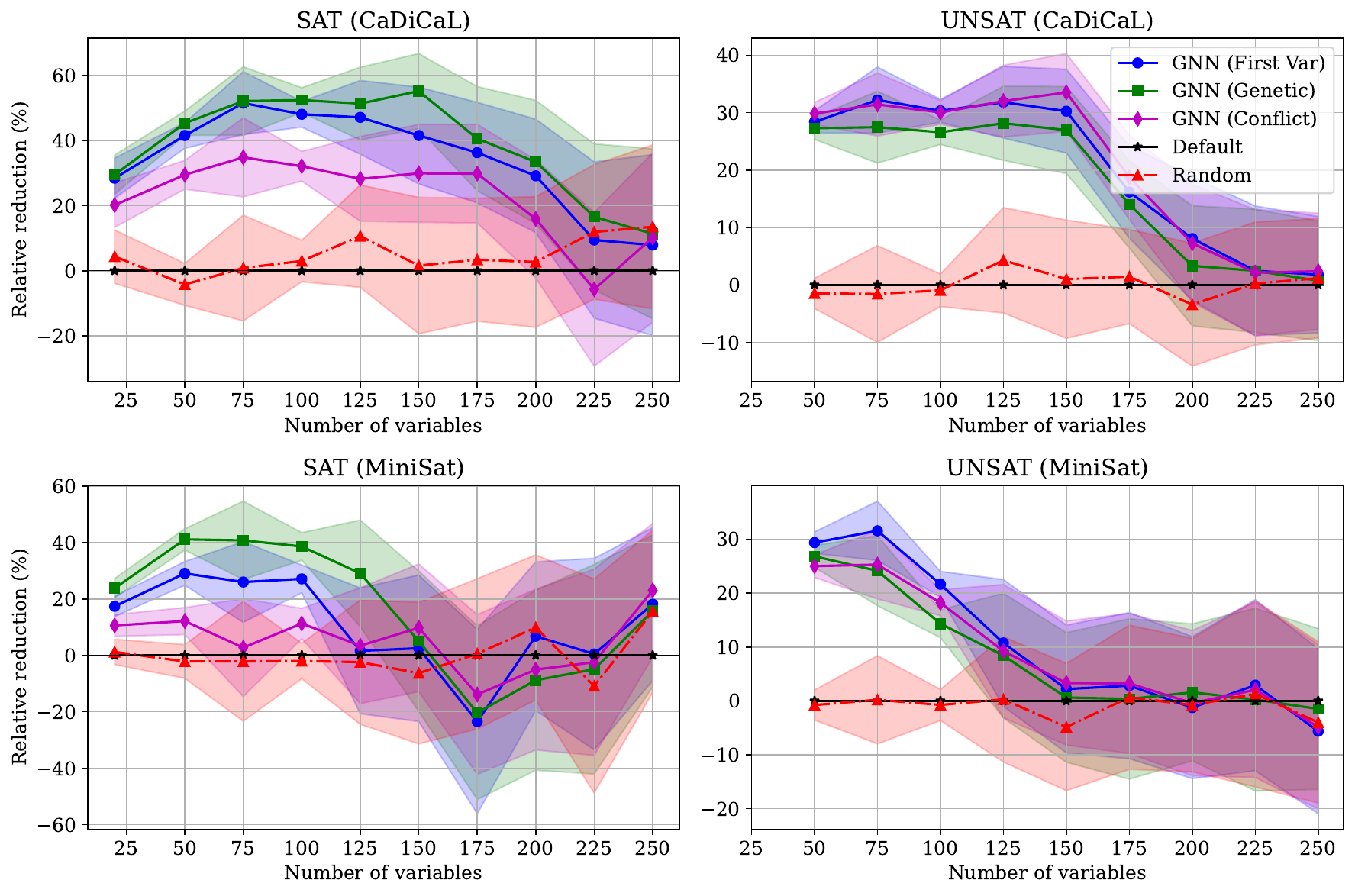}

    \caption{Mean reduction in the number of propagations relative to the default variable order on SATLIB uniform random 3-CNF benchmarks. Top: CaDiCaL on SAT and UNSAT instances, respectively. Bottom: MiniSat on SAT and UNSAT instances, respectively. Shaded regions denote a 95\% confidence interval across the dataset.}
    \label{fig:solver_results}
\end{figure}

Figure \ref{fig:solver_results} displays the relative reduction in the solving time compared to the default order for the  (branching in order of variable ID). We see a large improvement using the GNN-enhanced solvers for instances with 200 variables or fewer, occasionally reducing the solving time by more than 50\%. The network, trained on instances with 20 to 40 variables, also generalizes to much larger instances, with noticeable speed-ups in CaDiCaL on instances up to 5-10x larger than those from training. However, we see a rapid decay in the performance gains for larger instances. 

Interestingly, the three labeling methods have similar performance, with conflict labeling being slightly worse for SAT instances. However, the three labeling methods are very different in terms of learnability: using CaDiCaL, conflict labeling has an average Spearman correlation of 0.37 between predicted and true order, while the same number is 0.33 and 0.062 for first variable and genetic variable. This suggests that the GNN is learning some underlying common feature across the labeling methods and that the suggested orders themselves, especially from the genetic labeling, are hard to predict. This also does not seem to change with higher magnitudes of data or time spent training, further suggesting that there is some underlying feature being estimated. 

Second, we investigate performance on the suite provided by G4SATBench \citep{li_g4satbench_2024}, which consists of a mix of random problems, combinatorial problems, and pseudo-industrial problems. It is therefore a suitable middle-ground between random 3-CNF SAT and industrial instances. We train on 1000 instances from each problem class and satisfiability in the suite under ``medium'' difficulty, resulting in a training set of 14\,000 problems. We then evaluate on 100 problems of each problem class and satisfiability from the suite under ``hard'' difficulty, resulting in 1400 test instances.

Figure \ref{fig:g4satbench} displays the results for G4SATBench for CaDiCaL. We leave the results for MiniSat in Appendix \ref{app:g4satbenchminisat} for brevity as they show no clear difference between the different methods. We observe that, similar to the random instances, there is a large propagation reduction (10\%-20\% for first variable and genetic labeling) for the easier instances. However, for the harder instances requiring more than $10^5$ propagations we do not find any distinct difference between the GNNs and the baselines. For these problems we experience an even larger divide among the labeling methods in predictive power. We have a Spearman correlation of 0.36, 0.15, and 0.006 for conflict, first variable, and genetic labeling, respectively. This can be explained by a higher number of variables, resulting in noisier labels from the first variable and genetic labeling. However, we see that training on the first variable and genetic labels still provides better performance than the conflict labels.

\begin{figure}[ht]
    \centering
    \includegraphics[width=0.8\linewidth]{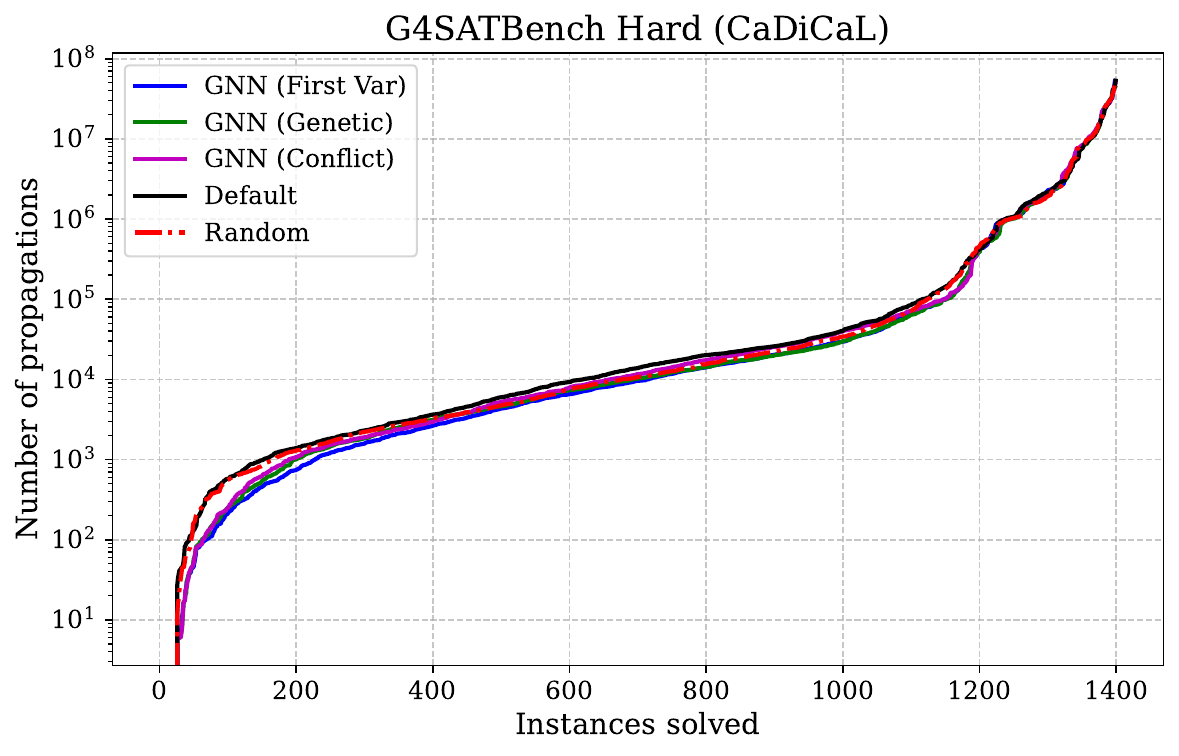}

    \caption{Number of propagations on the hard G4SATBench instances. Speedups of 10\%-20\% until roughly $10^5$ propagations.}
    \label{fig:g4satbench}
\end{figure}

\begin{figure}[ht]
    \centering
    \includegraphics[width=\linewidth]{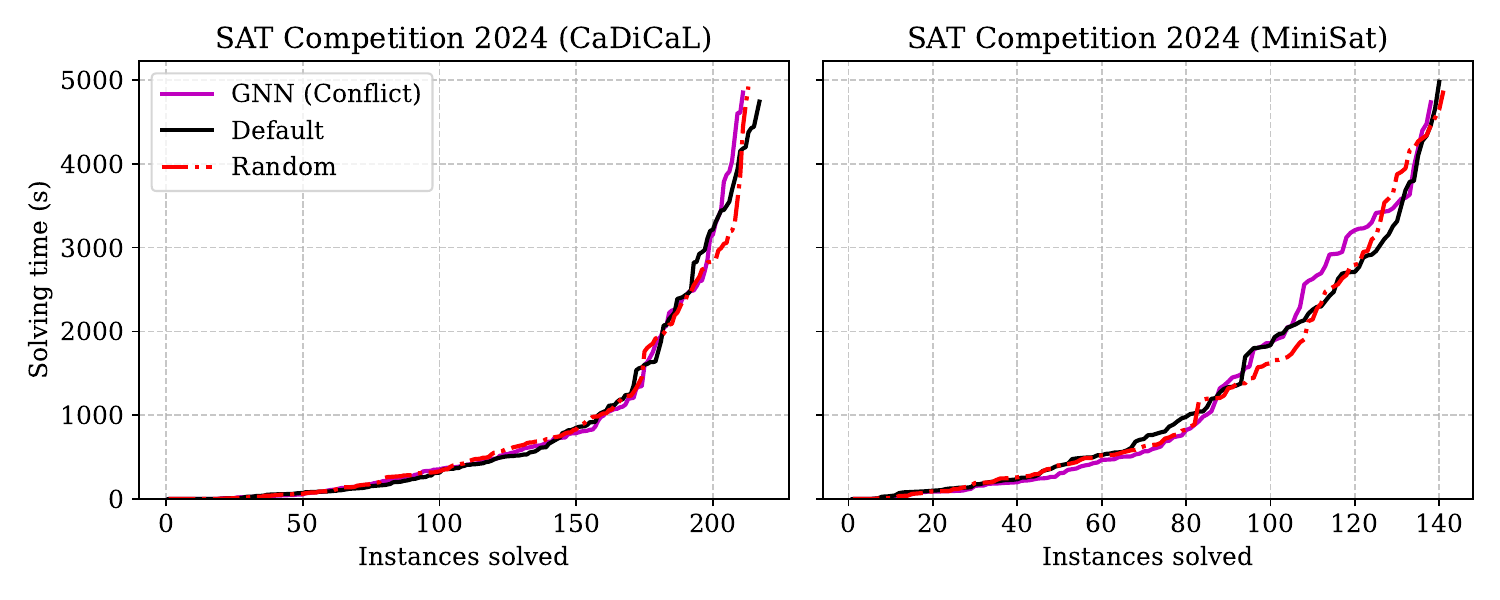}
    \caption{Solving time for CaDiCaL and MiniSat on the SAT competition 2024 main track.}
    \label{fig:satcomp_2024}
\end{figure}

Finally, we investigate the performance on industrial instances. Here we train on instances from the DataBack dataset \citep{wang_neuroback_2024-1}, consisting of various, mostly generated instances, before finetuning on the SAT competition main tracks from 2015 to 2023. We then evaluate the performance on the 2024 SAT competition's main track. Since the instances in DataBack and for the SAT competition are significantly harder to solve, it becomes computationally infeasible to use first variable and genetic labeling (since they require solving an instance many times to label). We therefore train a single GNN using conflict labeling, as seen in Figure \ref{fig:satcomp_2024}. Here we do not see any significant difference compared to the baselines. 

We theorize that the reason for the middling performance on harder instances for all experiments is twofold. First, we believe that the longer solving time simply leads to the order suggested by the GNN being overwritten by the solver's own heuristics, thus voiding its contribution. This would also explain MiniSat’s inferior performance, since our suggested order is imposed only through VSIDS activities, which are quickly overwritten without the VMTF queue present in CaDiCaL.
Secondly, we believe that the complexity of these problems make it harder for the GNN to reliably predict a good initial order. To investigate the former, we conduct a preliminary experiment of reminding the solver of the suggested order seen in Appendix \ref{app:reminding}.




\section{Conclusion and Future Work}
We investigate predicting initial branching order as a preprocessing step to SAT solvers. Our experiments show that there are gains to be made by predicting a good initial branching order. Furthermore, we suggest three different ways to label such branching orders and show that their performance is similar when used as labels to a GNN. Lastly, we show that a GNN trained on such labels can enhance the performance of state-of-the-art models on generated and pseudo-industrial instances, although with a decay in improvement as the number of variables increases. This gives a strong indication that GNNs are able to exploit the structure of SAT problems to enhance the initial branching order of the state-of-the-art solvers, although as problems become harder to solve, their effects seems to diminish.

\section{Acknowledgements}
The computations were enabled by resources provided by the National Academic Infrastructure for Supercomputing in Sweden (NAISS), partially funded by the Swedish Research Council through grant agreement no. 2022-06725. This work was supported in part by Swedish Research Council Distinguished
Professor Grant 2017-01078, Knut and Alice Wallenberg Foundation Wallenberg Scholar Grant, and Swedish Strategic Research Foundation FUSS SUCCESS Grant. 


\bibliography{references}
\bibliographystyle{iclr2026_conference}








\appendix

\section{Labeling Pseudocode}
\label{app:labelapp}

The following pseudocode describes the labeling algorithms in as much detail as needed to implement them. For genetic labeling, we also let the random swaps decay such that the distances of the random swaps are halved each generation.

\begin{algorithm}[H]
\caption{First variable labeling}
\label{alg:first-variable}
\KwIn{SAT instance $\phi$, solver $f_{solve}$, number of trials per variable $k$}
\KwOut{Variable order $\pi_{label}$}

$X \gets$ variables in $\phi$ \;
\ForEach{$x_i \in X$}{
    score[$x_i$] $\gets 0$ \;
    \For{$j \gets 1$ \KwTo $k$}{
        $\pi_j \sim \mathcal{U}(\text{all permutations of } X \setminus \{x_i\})$ \;
        order $\gets (x_i\ ||\ \pi_j)$ \;
        score[$x_i$] $\gets$ score[$x_i$] $+ f_{solve}(\phi, \text{order})$ \;
    }
    score[$x_i$] $\gets$ score[$x_i$] $/ k$ \;
}
$\pi_{label} \gets$ variables in $X$ sorted by score \;
\Return $\pi_{label}$ \;
\end{algorithm}

\begin{algorithm}[H]
\caption{Genetic labeling}
\label{alg:genetic}
\KwIn{SAT instance $\phi$, solver $f_{solve}$, population size $k$, generations $m$}
\KwOut{Variable order $\pi_{label}$}

$X \gets$ variables in $\phi$ \;
\For{$j \gets 1$ \KwTo $k$}{
    $\pi_j \sim \mathcal{U}(\text{all permutations of } X)$ \;
    score[$\pi_j$] $\gets f_{solve}(\phi, \pi_j)$ \;
}
$\pi_{best} \gets \arg\min_{\pi_j} \text{score}[\pi_j]$ \;

\For{$g \gets 1$ \KwTo $m$}{
    $l \gets |\pi_{best}|/2$ \;
    \For{$j \gets 1$ \KwTo $k$}{
        $\pi'_j \gets$ Apply $|\pi_{best}|$ random swaps to $\pi_{best}$ with maximum swap length $l$\;
        score[$\pi'_j$] $\gets f_{solve}(\phi, \pi'_j)$ \;
    }
    $\pi_{best} \gets \arg\min_{\pi'_j} \text{score}[\pi'_j]$ \;
    $l \gets l/2$ \;
}
\Return $\pi_{best}$ as $\pi_{label}$ \;
\end{algorithm}



\section{Computational Setup}
\label{app:setup}
The experiments are performed on a single NVIDIA A100 GPU when training the graph neural network and using an AMD EPYC 7742 64-Core Processor CPU when labeling and testing. Labeling the 10K instances by solving each instance 100 times takes less than a minute using all 64 cores. Training for 10 epochs also takes less than a minute.

\section{Inference Time}
\label{app:inference}

\begin{figure}[ht]
    \centering
    \begin{subfigure}[t]{0.48\linewidth}
        \centering
        \includegraphics[width=\linewidth]{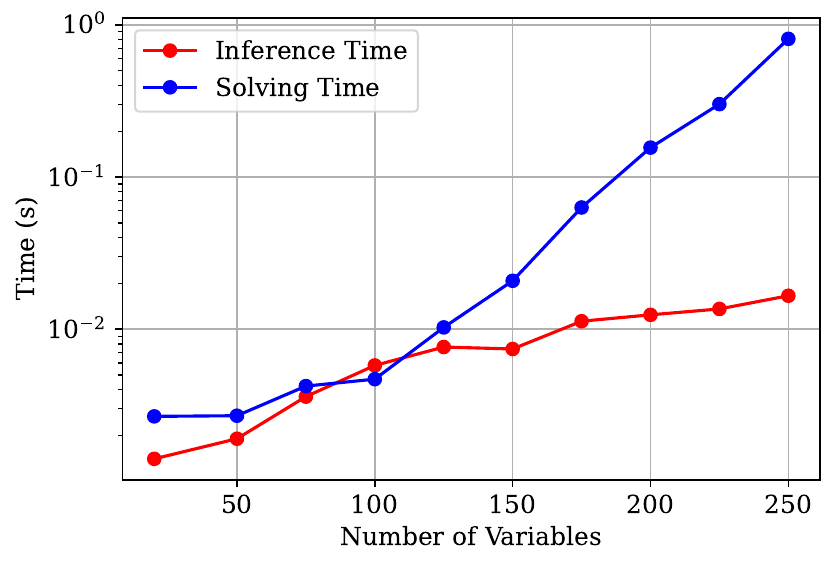}
        \caption{SAT instances}
        \label{fig:sat_inference}
    \end{subfigure}
    \hfill
    \begin{subfigure}[t]{0.48\linewidth}
        \centering
        \includegraphics[width=\linewidth]{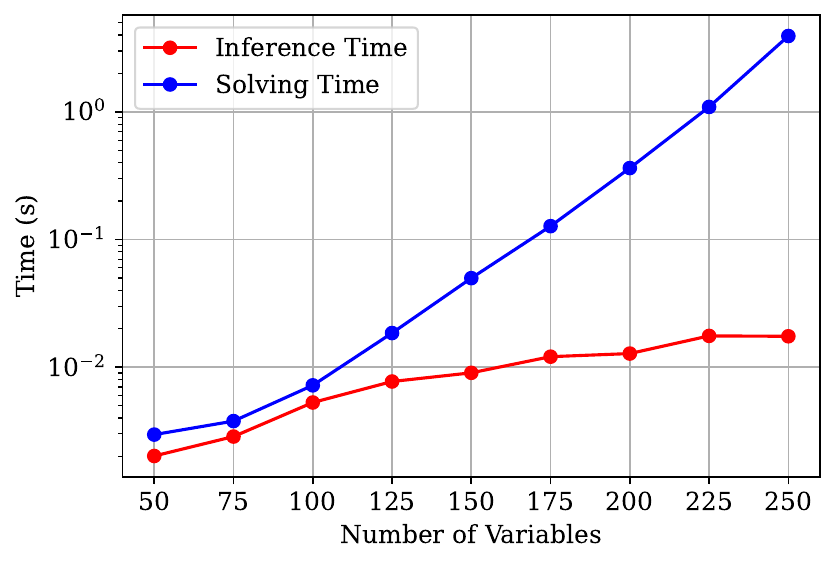}
        \caption{UNSAT instances}
        \label{fig:unsat_inference}
    \end{subfigure}
    \caption{Comparison of inference time and solving time using CaDiCaL (default order) when solving on CPU only for SAT (a) and UNSAT (b) instances. Note the logarithmic y-axis.}
    \label{fig:inference_time}
\end{figure}

For a fair comparison, we compare the inference time of the GNN on CPU to the solving time of CaDiCaL in Figure \ref{fig:inference_time} to see for which random 3-CNF instances the cost is negligible. We see, as expected, that inference time exhibits a linear pattern, while solving time exhibits an exponential pattern. Combining this with the results in Figure \ref{fig:solver_results} shows that the increased cost for inference is worth it after roughly 150 variables. The same concern is not needed when doing inference on GPU, as this is much faster.

\section{Model Hyperparameters}
\label{app:hyperparams}

During training, we use the hyperparameters shown in Table \ref{tab:hyperparams}. 

\begin{table}[h!]
\centering
\caption{Model hyperparameters}
\begin{tabular}{>{\raggedright\arraybackslash}p{4cm} p{6cm}}
\toprule
\textbf{Hyperparameter} & \textbf{Value} \\
\midrule
Model Architecture & Graph Transformer from NeuroBack \citep{wang_neuroback_2024-1} \\
Optimizer & AdamW \citep{loshchilov_decoupled_2018} \\
Adam $\beta_1$ & 0.9 \\
Adam $\beta_2$ & 0.999 \\
Learning Rate & 0.001 \\
Epochs & 10 \\
Batch Size & 128 \\
GSA Blocks & 3 \\
LSA Blocks & 3 \\
Dropout Rate & 0.2 \\
Weight Decay & 0.01 \\
Gradient Clipping & 1.0 \\
Loss Function & LambdaRank \\
\bottomrule
\end{tabular}
\label{tab:hyperparams}
\end{table}

\section{MiniSat on G4SATBench}
\label{app:g4satbenchminisat}

\begin{figure}[ht]
    \centering
    \includegraphics[width=0.8\linewidth]{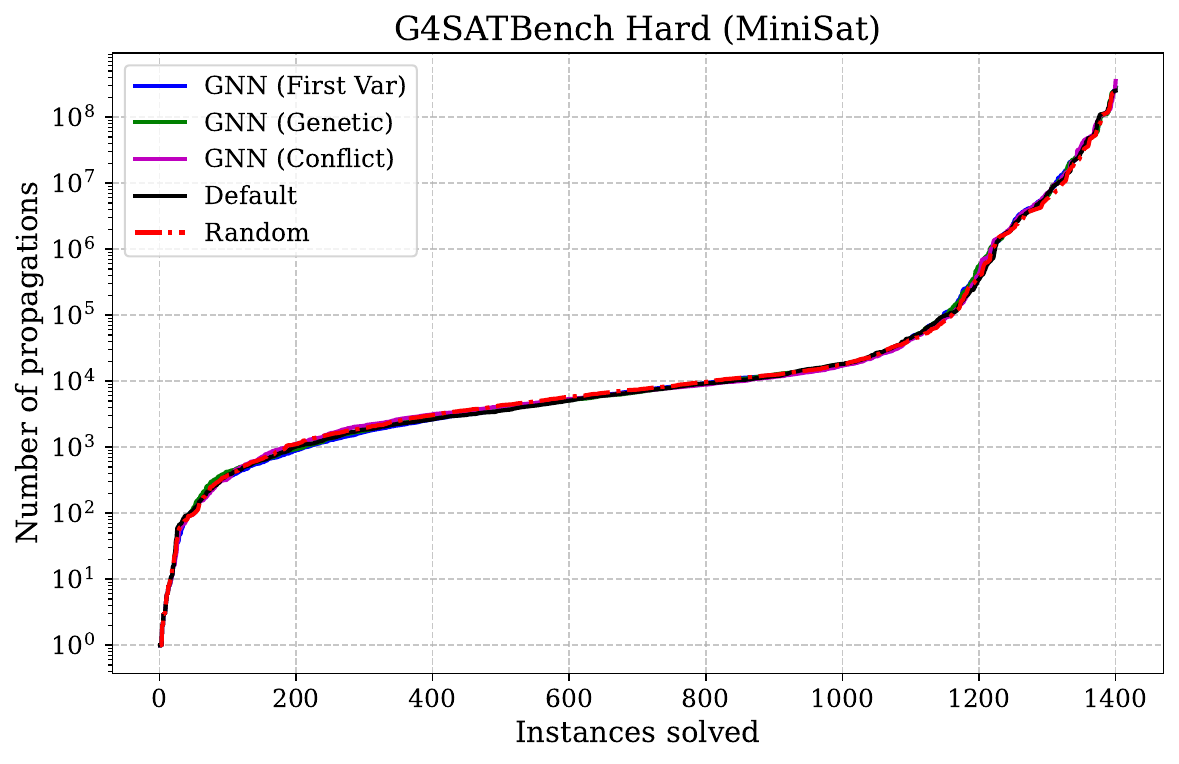}

    \caption{Number of propagations on the hard G4SATBench instances. No distinct speedup.}
    \label{fig:g4satbenchminisat}
\end{figure}

See Figure \ref{fig:g4satbenchminisat}.

\section{Reminding of the Suggested Order}
\label{app:reminding}

We investigate if reminding the solver of the suggested branching order can help performance for instances that take longer to solve. We do this by adding the suggested order to the VSIDS activities every restart (roughly every 1000 conflicts) by updating the activities of variable $i$ through 

$$ activities(i) \leftarrow remind\ factor\times max(activities) \times decay\ factor^{-R_{GNN}(i)}$$

where activities is the VSIDS activities and the decay factor is $0.95$. $R_{GNN}(i)$ is the rank of the variable in the suggested order. 

\begin{figure}[h]
    \centering
    \includegraphics[width=\linewidth]{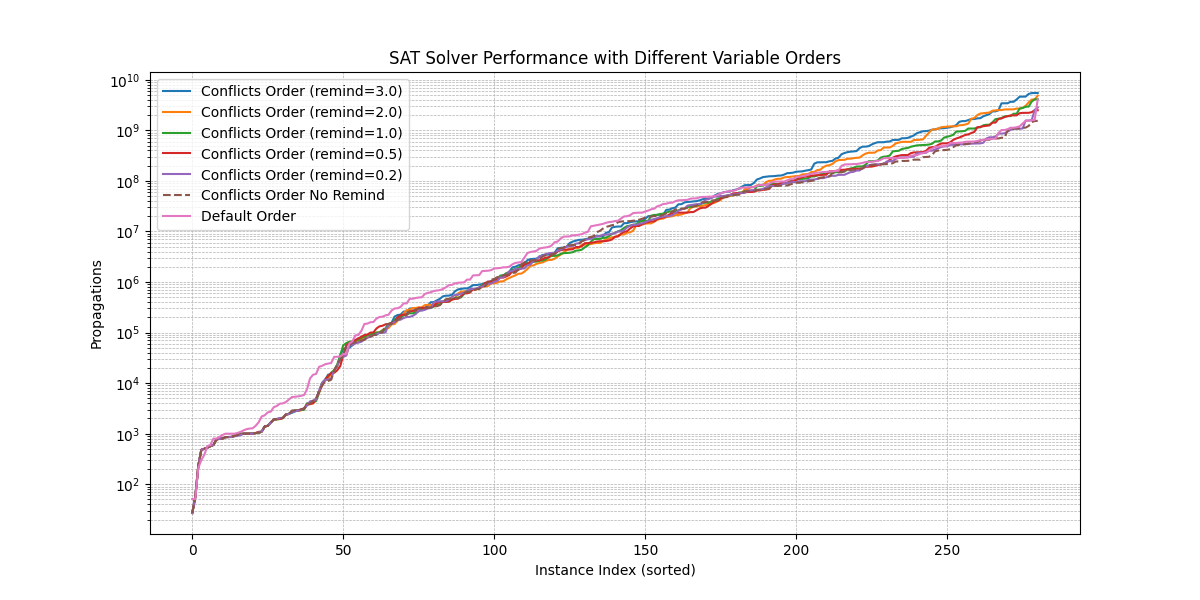}
    \caption{Number of propagations for different reminding factors. Only the problems that all seven initializations solve are plotted.}
    \label{fig:reminding}
\end{figure}

We evaluate this using MiniSat \citep{sorensson_minisat_2005} on SAT competitions 2015 to 2024 with a timeout of 5000, where the suggested order is the conflict label for different remind factors. Figure \ref{fig:reminding} shows the results. For larger instances, we see that not reminding is the fastest option, thus showing that the naive solution of reminding every restart does not seem to alleviate our issue of not speeding up difficult-to-solve instances. The poor performance of reminding can also be motivated by our continuously overwriting the vital VSIDS activities, thus making it harder for the solver to explore the problem properly. 

\end{document}